\documentclass[10pt,twocolumn,letterpaper]{article}

\usepackage{cvpr}              % To produce the CAMERA-READY version
\usepackage{times}
\usepackage{epsfig}
\usepackage{graphicx}
\usepackage{amsmath}
\usepackage{amssymb}

\usepackage[caption = false]{subfig}
\usepackage{comment}
\usepackage{color}

\usepackage{dsfont}
\usepackage{multirow}
\usepackage{cite}

\usepackage{verbatim}
\usepackage{slashbox}
\usepackage{array}
\usepackage{booktabs, caption, makecell}

\usepackage{algorithm}
\usepackage{algpseudocode}%

\usepackage[pagebackref=true,breaklinks=true,letterpaper=true,colorlinks,bookmarks=false]{hyperref}

\newcolumntype{L}[1]{>{\raggedright\arraybackslash}m{#1}}
\newcolumntype{C}[1]{>{\centering\arraybackslash}m{#1}}
\newcolumntype{R}[1]{>{\raggedleft\arraybackslash}m{#1}}
\newcolumntype{+}{>{\global\let\currentrowstyle\relax}}
\newcolumntype{^}{>{\currentrowstyle}}

\newcommand{\bfo}{\ensuremath{{\mathbf{o}}}}

\newcommand{\bfr}{\ensuremath{{\mathbf{r}}}}
\newcommand{\bfx}{\ensuremath{{\mathbf{x}}}}

\newcommand{\bfc}{\ensuremath{{\mathbf{c}}}}
\newcommand{\bfu}{\ensuremath{{\mathbf{u}}}}
\newcommand{\bfv}{\ensuremath{{\mathbf{v}}}}
\newcommand{\bfU}{\ensuremath{{\mathbf{U}}}}

\newcommand{\bfA}{\ensuremath{{\mathbf{A}}}}
\newcommand{\bfV}{\ensuremath{{\mathbf{V}}}}

\usepackage[capitalize]{cleveref}
\crefname{section}{Sec.}{Secs.}
\Crefname{section}{Section}{Sections}
\Crefname{table}{Table}{Tables}
\crefname{table}{Tab.}{Tabs.}

\def\cvprPaperID{6936} % *** Enter the CVPR Paper ID here
\def\confName{CVPR}
\def\confYear{2022}

\begin{document}

% \title{Eigenshapes: A Data-Driven Shape Descriptors and Its Applications}
\title{Applying Eigencontours to PolarMask-Based Instance Segmentation}

\author{Wonhui Park\\
Korea University\\
{\tt\small whpark@mcl.korea.ac.kr}
\and
Dongkwon Jin\\
Korea University\\
{\tt\small dongkwonjin@mcl.korea.ac.kr}
\and
Chang-Su Kim\\
Korea University\\
{\tt\small changsukim@korea.ac.kr}
}
\maketitle

%%%%%%%%% ABSTRACT
\begin{abstract}
    Eigencontours are the first data-driven contour descriptors based on singular value decomposition. Based on the implementation of ESE-Seg, eigencontours were applied to the instance segmentation task successfully. In this report, we incorporate eigencontours into the PolarMask network for instance segmentation. Experimental results demonstrate that the proposed algorithm yields better results than PolarMask on two instance segmentation datasets of COCO2017 and SBD. Also, we analyze the characteristics of eigencontours qualitatively. Our codes are available at \href{https://github.com/dnjs3594/Eigencontours}{https://github.com/dnjs3594/Eigencontours}.
\end{abstract}

%%%%%%%%% BODY TEXT
\section{Introduction}

Instance segmentation is a task to delineate the shape of each distinct object in an image. Most instance segmentation techniques are based on the framework of object detection, such as Mask R-CNN \cite{he2017} and its variants \cite{liu2018path,chen2019,fang2021}. They detect the bounding box of each instance and then classify whether each pixel within the box belongs to the instance or not.   However, despite yielding promising results, they often demand high computational costs due to the pixelwise classification.

To alleviate the computational complexity, contour-based techniques \cite{xu2019,xie2020,park2022} also have been developed. These techniques reformulate the pixelwise classification task as the boundary regression task of an object. To this end, they represent object boundaries with contour descriptors. For example, in ESE-Seg \cite{xu2019}, the boundary of an object is represented by polynomial fitting coefficients. Also, PolarMask \cite{xie2020} describes an object boundary in polar coordinates using centroidal profiles. Then, these techniques localize objects by predicting the contour description vectors only, which is computationally more efficient than the pixelwise classification. However, they may fail to reconstruct object boundaries accurately.

Recently, in \cite{park2022}, we proposed eigencontours, which are data-driven contour descriptors obtained by the low-rank approximation of all object boundaries in a training set through singular value decomposition (SVD). Experimental results showed that an object boundary can be described faithfully by a linear combination of a small number of eigencontours, and that eigencontours are more effective and more efficient than the conventional contour descriptors \cite{xu2019,xie2020}. Also, it was shown that, based on the framework of ESE-Seg, eigencontours improve the instance segmentation performance on the SBD dataset \cite{hariharan2011}.

In this report, based on the framework of PolarMask \cite{xie2020}, we apply eigencontours to the instance segmentation task. First, we describe how to construct eigencontours in detail. Then, we design a network by modifying the PolarMask network. It is shown experimentally that the proposed network outperforms PolarMask on the COCO2017 \cite{lin2014} and SBD \cite{hariharan2011} datasets. We also discuss the strengths and weaknesses of eigencontour descriptors qualitatively.

\begin{figure}[t]
  \centering
  \includegraphics[width=1\linewidth]{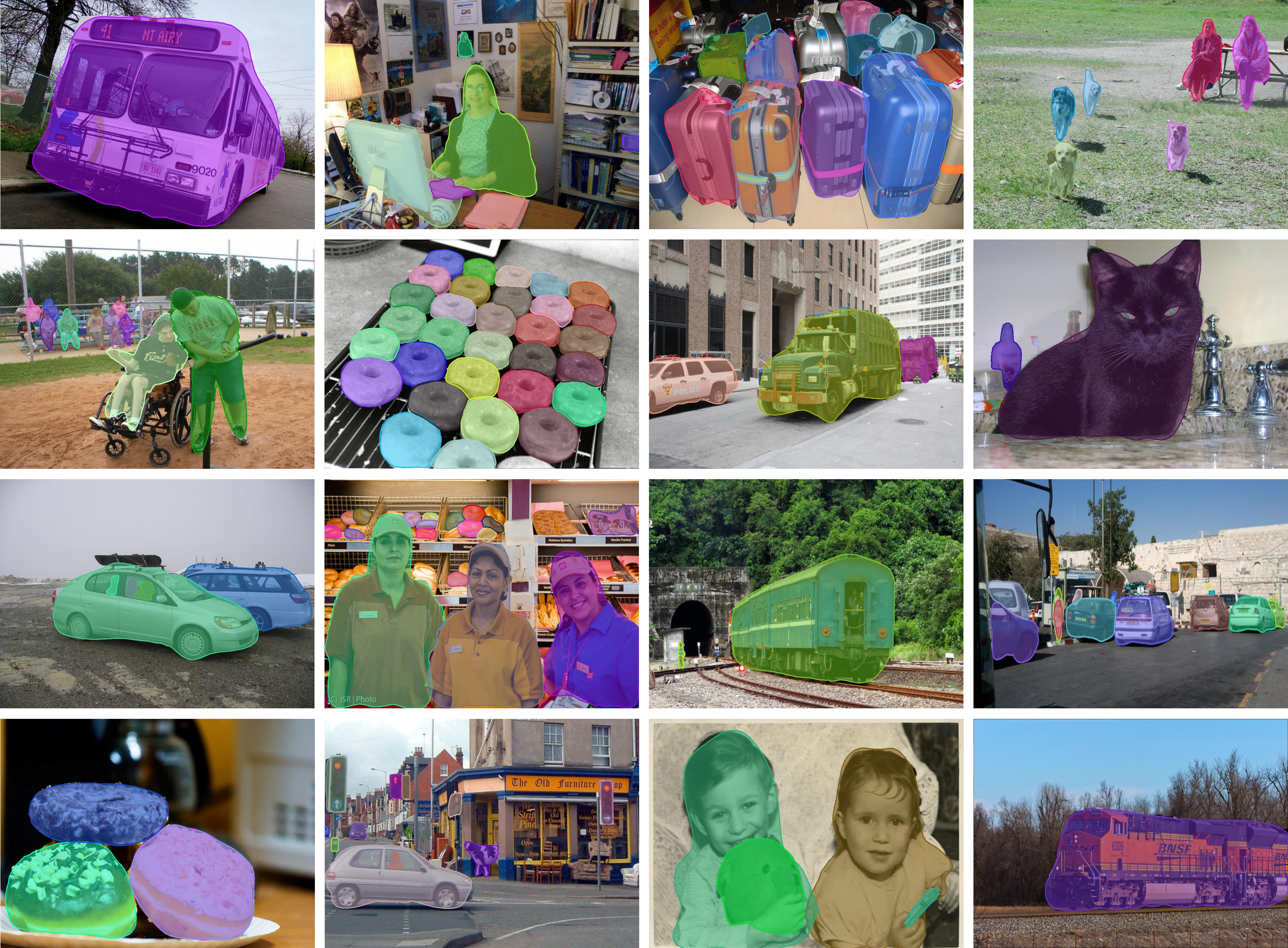}
  \caption{Instance segmentation results of the proposed algorithm on the COCO2017 validation dataset \cite{lin2014}.}
  \label{fig:Intro_fig}
  % \vspace*{0.3cm}
\end{figure}

\section{Eigencontour Representation}
Let us describe how to represent object boundaries in the eigencontour space \cite{park2022}. First, a star-convex contour is generated from each object boundary in a training set. Second, a star-convex contour matrix is constructed to include all the star-convex contours in the training set as its column vectors. Third, the star-convex contour matrix is decomposed into eigencontours via SVD. Last, an object boundary is represented by a linear combination of the first $M$ eigencontours.

\label{ssec:preprocessing}

\vspace*{0.1cm}
\noindent\textbf{Star-convex contour generation:}
In \cite{park2022}, object boundaries are first simplified to star-convex contours to strike a balance between the accuracy and the simplicity of contour representation. To this end, centroidal profiles \cite{davies2004} are employed. Given an object shape, a center point $\bfo=(x,y)$ is defined, which can be determined as the inner-center \cite{xu2019} or the centroid \cite{xie2020}. Then, the boundary is described using polar coordinates $(r_i, \theta_i)$, $i=1, 2, \ldots, N$. The angular coordinates $\theta_i$ are sampled uniformly, so only the radial coordinates are used to represent the contour
\begin{equation}
\bfr=[r_1, r_2, \ldots, r_N]^\top
\label{eq:contour_def}
\end{equation}
where $r_i$ is the distance of the farthest object point from the center along the $\theta_i$-axis. Thus, a star-convex contour is expressed by $\bfr$.

\vspace*{0.1cm}
\noindent\textbf{Eigencontour construction:}
Let ${\mathbf A}=[\bfr_1, \bfr_2, \cdots, \bfr_L]$ be a star-convex contour matrix, containing $L$ star-convex contours of objects in a training set. Then, by performing SVD, the contour matrix $\mathbf A$ is decomposed into
\begin{equation}\label{eq:svd}
    \textstyle
    \mathbf{A} = \mathbf{U} \mathbf{\Sigma} \mathbf{V}^\top
\end{equation}
where $\mathbf{U} = [\bfu_1, \cdots, \bfu_N]$ and $\bfV = [\bfv_1, \cdots, \bfv_L]$ are orthogonal matrices and $\mathbf{\Sigma}$ is a diagonal matrix, composed of singular values $\sigma_1 \geq \sigma_2 \geq \cdots \geq \sigma_r > 0$. Here, $r$ is the rank of $\mathbf A$. Then, the best rank-$M$ approximation of $\mathbf A$ \cite{y2015SVD} is given by
\begin{equation}\label{eq:rank-m}
    \bfA_M  = [\tilde{\bfr}_1, \cdots, \tilde{\bfr}_L] = \sigma_{1}{\mathbf u}_{1}{\mathbf v}^\top_{1} + \cdots + \sigma_{M}{\mathbf u}_{M}{\mathbf v}^\top_{M}.
\end{equation}
The first $M$ left singular vectors $\bfu_1, \cdots, \bfu_M$ are referred to as \textit{eigencontours}. Also, the space spanned by $\{\bfu_1, \cdots, \bfu_M\}$ is called the \textit{eigencontour space}.

\vspace*{0.1cm}
\noindent\textbf{Contour representation:}
We can approximate a contour  $\bfr$ by a linear combination of the first $M$ eigencontours,
\begin{equation} \label{eq:x_approx}
\tilde{\bfr} = \bfU_M \bfc = [\bfu_1, \cdots, \bfu_M] \bfc
\end{equation}
where the coefficient vector $\bfc$ is given by
\begin{equation}\label{eq:forward}
    \bfc = \bfU_M^\top \bfr.
\end{equation}
This means that an $N$-dimensional contour $\bfr$ is optimally approximated by an $M$-dimensional vector $\bfc$ in the eigencontour space, where $M < N$. Also, the approximate $\tilde{\bfr}$ can be reconstructed from $\bfc$ via \eqref{eq:x_approx}. {\bf Algorithm 1} summarizes how to construct the eigencontour space.

\begin{algorithm}[t]
\caption{Eigencontour construction}
    {\bf Input:} Set of training objects $\{\bfx_1, \cdots, \bfx_L\}$, $N=$ \# of star-convex contour points, $M=$ \# of eigencontours.
    \begin{algorithmic}[1]
        \State Generate $N$-dimensional star-convex contour vectors $\{\bfr_{1}, \cdots, \bfr_{L}\}$ and center points $\{\bfo_{1}, \cdots, \bfo_{L}\}$ for all training objects
        \State Construct the star-convex contour matrix ${\mathbf A}$ and apply SVD in \eqref{eq:svd}
        \State Perform the best rank-$M$ approximation and obtain the first $M$ eigencontours via \eqref{eq:rank-m}
        \State Approximate each contour $\bfr_i$ to $\bfc_i$ via \eqref{eq:x_approx}
    \end{algorithmic}
    {\bf Output:} Set of coefficients $\{\bfc_{1}, \cdots, \bfc_{L}\}$ and center points $\{\bfo_{1}, \cdots, \bfo_{L}\}$
    \label{alg:eigencontour algorithm}
\end{algorithm}

\begin{figure*}[t]
  \centering
  \includegraphics[width=1\linewidth]{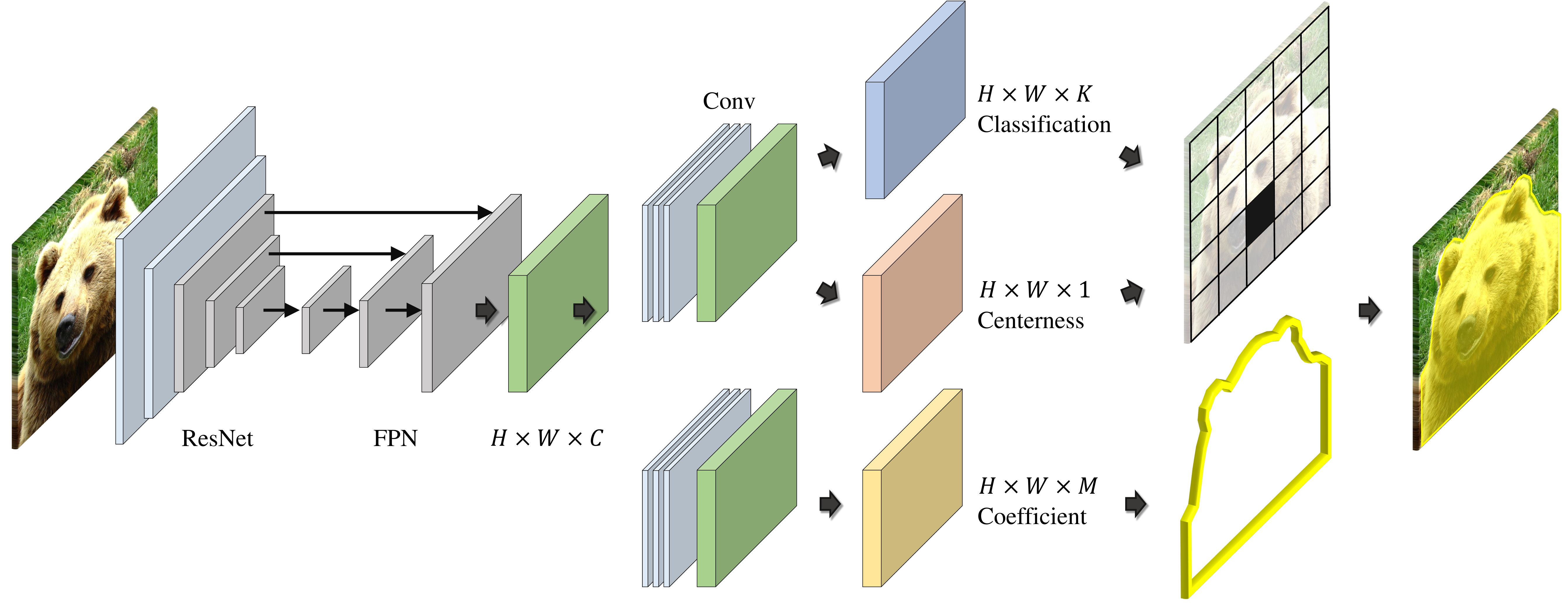}
  \setlength{\abovecaptionskip}{0.1cm}
  \caption{The architecture of the proposed instance segmentation network based on PolarMask.}
  \label{fig:overview_fig}
\end{figure*}

\section{Instance Segmentation Using Eigencontours}
We apply eigencontours into instance segmentation based on the implementation of PolarMask \cite{xie2020}.

\vspace*{0.1cm}
\noindent\textbf{Architecture:}
For instance segmentation, we design a network based on PolarMask \cite{xie2020}. Figure~\ref{fig:overview_fig} shows the structure of the proposed network, which is composed of a single encoder and three decoders. Given an image, the encoder extracts a convolutional feature map of size $H\times W\times C$. Then, from the feature map, the three decoders generate output maps, respectively: $P \in \mathbb{R}^{H\times W\times K}$, $O \in \mathbb{R}^{H\times W\times 1}$, and $R \in \mathbb{R}^{H\times W\times M}$. Each element in $P$ is a $K$-dimensional vector, indicating whether the corresponding pixel belongs to an object in one of the $K$ predefined categories. Each score in $O$ represents the probability that the corresponding pixel is the center point of an object. Each element in $R$ is a $M$-dimensional vector, which, in the eigencontour space,  represents the contour of the object centered on the corresponding pixel.
As in \cite{xie2020}, the encoder is based on the feature pyramid network with the ResNet50 backbone \cite{he2016deep}. Also, each decoder is implemented by a series of 2D convolution layers.

\vspace*{0.1cm}
\noindent\textbf{Coefficient regression:}
In PolarMask, an object boundary in an image is represented by a star-convex contour. To detect the object boundary, it should regress $N$ variables in $\bfr$ in \eqref{eq:contour_def}. On the other hand, in this work, an object contour is represented by a linear combination of the $M$ eigencontours. Thus, the proposed network needs to regress only the $M$ coefficients of $\bfc$ in \eqref{eq:forward} in the eigencontour space.

\vspace*{0.1cm}
\noindent\textbf{Postprocessing:}
From the output maps, we extract the segmentation masks of objects also based on the postprocessing scheme of PolarMask. For each pixel $i$, we first compute a confidence score by multiplying the score $O_i$ with the maximum probability in $P_i$. We then filter out the pixels with confidence scores lower than 0.05. Then, we perform non-maximum suppression. Specifically, we iteratively select the pixel with the maximum confidence score and remove the pixels whose decoded masks overlap with the mask of the selected pixel. Finally, we declare the segmentation masks of the selected pixels as the final results. Notice that, to get a segmentation mask, we reconstruct the contour vector by linearly combining the eigencontours using the coefficients in $R_i$. Then, we convert the contour vector to contour points using the uniformly sampled angular coordinates.

%-------------------------------------------------------------------------
\section{Experiments}

\subsection{Datasets}
\label{ssec:dataset}

We conduct experiments on the COCO2017 \cite{lin2014} and SBD \cite{hariharan2011} datasets. COCO2017 consists of 118K training and 5K validation images, and SBD contains 5,623 training and 5,732 validation images. For COCO2017 and SBD, instance segmentation masks are provided for objects in $K=80$ and 20 categories, respectively.

\subsection{Training details}
We use the stochastic gradient descent (SGD) optimizer and train the network for 12 epochs. We set the initial learning rate to 0.01 and decrease it by a factor of 0.1 after 8 and 11 epochs, respectively. We set the weight decay to 0.0001 and the momentum to 0.9. The batch size is 4, and the size of an input image is $768 \times 1280$. To train the network, we minimize the loss
$$L = L_{\rm cls} + L_{\rm cen} + L_{\rm coeff}$$
where $L_{\rm cls}$ is the focal loss \cite{lin2017focal} between a predicted map $P_i$ and the ground-truth $\bar{P}_i$, $L_{\rm cen}$ is the binary cross-entropy loss between a predicted map $O_i$ and ground-truth $\bar{O}_i$, and $L_{\rm cls}$ is the polar IoU loss \cite{xie2020} between a predicted map $R_i$ and the ground-truth $\bar{R}_i$.

\begin{table}[t]\centering
    \renewcommand{\arraystretch}{0.9}
    \caption
    {
        Comparison of $\text{AP}_{50}$, $\text{AP}_{75}$, and $\text{AP}$ performances on the COCO2017 validation dataset.
    }
    \vspace*{-0.1cm}
    \resizebox{0.95\linewidth}{!}{
    \begin{tabular}[t]{+L{3.0cm}^C{1.6cm}^C{1.6cm}^C{1.6cm}}
    \toprule
    & $\text{AP}$ & $\text{AP}_{50}$ & $\text{AP}_{75}$ \\
    \midrule
         PolarMask \cite{xie2020}      & 29.0 & 48.9 & 29.8\\
         Proposed                  & \bf{29.6} & \bf{49.9} & \bf{30.4}\\
    \bottomrule
    \end{tabular}}
    \vspace{0.2cm}
    \label{table:result_coco}
\end{table}

\begin{table}[t]\centering
    \renewcommand{\arraystretch}{0.9}
    \caption
    {
        Comparison of $\text{AP}_{50}$, $\text{AP}_{75}$, and $\text{AP}$ performances on the SBD validation dataset.
    }
    \vspace*{-0.1cm}
    \resizebox{0.95\linewidth}{!}{
    \begin{tabular}[t]{+L{3.0cm}^C{1.6cm}^C{1.6cm}^C{1.6cm}}
    \toprule
    & $\text{AP}$ & $\text{AP}_{50}$ & $\text{AP}_{75}$ \\
    \midrule
         PolarMask \cite{xie2020}      & 27.7 & 50.6 & 25.7 \\
         Proposed                  & \bf{30.7} & \bf{54.9} & \bf{29.6} \\
    \bottomrule
    \end{tabular}}
    \vspace{0.2cm}
    \label{table:result_sbd}
\end{table}

\begin{figure*}[t]
  \centering
  \includegraphics[width=1\linewidth]{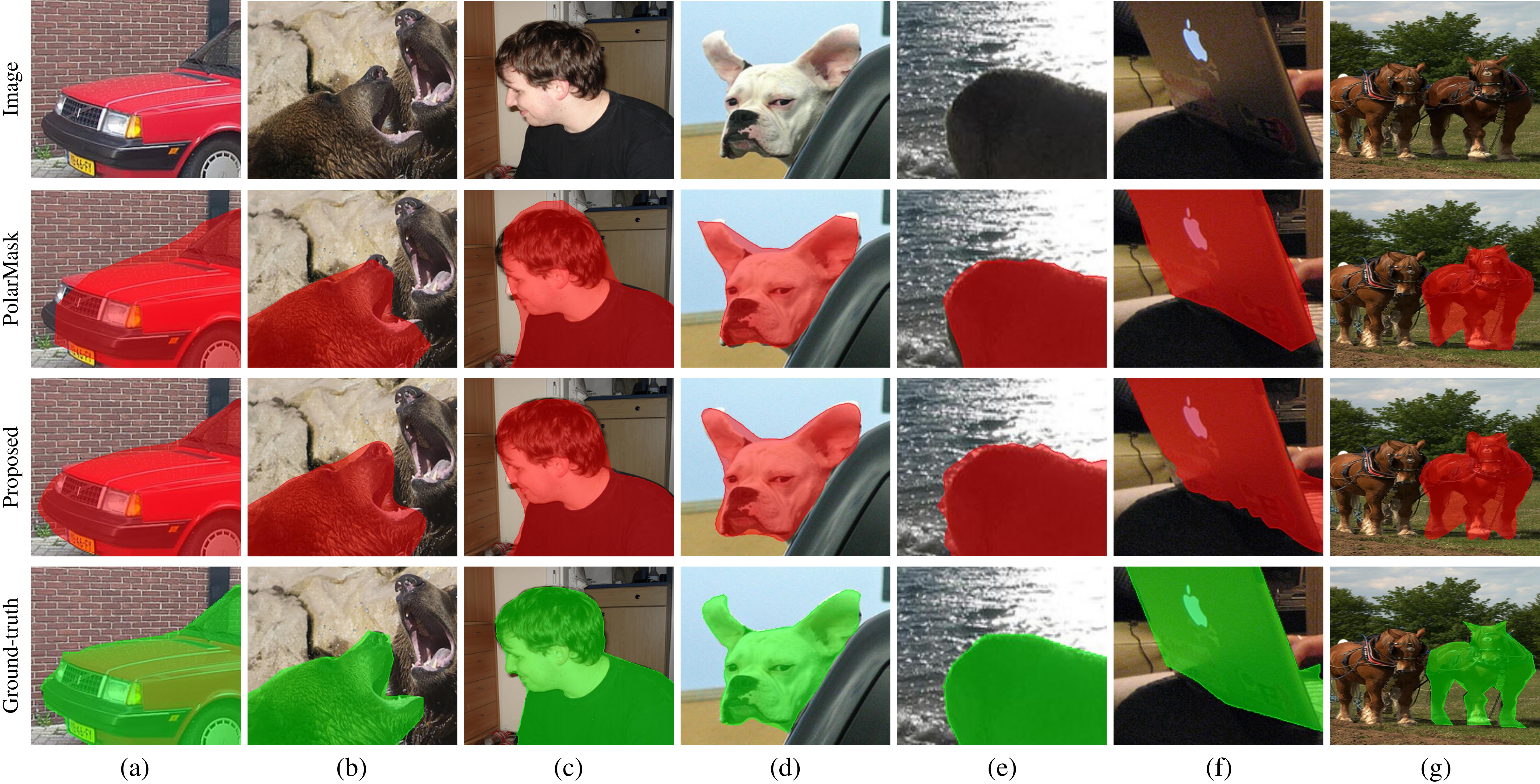}
  \setlength{\abovecaptionskip}{0.1cm}
  \caption{Qualitative comparison of instance segmentation results on the COCO2017 and SBD datasets.}  \label{fig:Result_inst_seg}
\end{figure*}

\subsection{Comparative assessment}
\label{ssec:assess}

\noindent\textbf{Quantitative comparison with PolarMask:}
We compare the instance segmentation results of the proposed network with those of PolarMask \cite{xie2020} on COCO2017 and SBD, by employing the same number $M$ of parameters. In PolarMask, an object contour is described by $M$ radial coordinates in a centroidal profile. On the other hand, in this work, the contour is represented by $M$ eigencontour coefficients.

Table \ref{table:result_coco} shows the results on the COCO2017 validation dataset at $M=36$. The average precision (AP) performance is measured, which averages AP scores across mask intersection-over-union (IoU) thresholds from 0.5 to 0.95 with an interval of 0.05. Also, we report $\text{AP}_{50}$ and $\text{AP}_{75}$ using thresholds 0.5 and 0.75, respectively. The proposed algorithm outperforms PolarMask in all three metrics. Table \ref{table:result_sbd} compares the results on the SBD validation dataset. The proposed algorithm yields better results than PolarMask by significant margins in terms of all three metrics.

The performance gaps between the proposed algorithm and PolarMask are marginal on COCO2017, whereas they are significant on SBD. For COCO2017, it is difficult to discover typical contour patterns due to the occlusions and the diversity of object shapes. Hence, eigencontour descriptors cannot represent the object boundaries properly. In contrast, for SBD, eigencontours describe object contours more reliably and thus more accurately, since the instances are classified into 20 categories only.

\vspace*{0.1cm}
\noindent\textbf{Qualitative comparison with PolarMask:}
Figure~\ref{fig:Result_inst_seg} compares the instance segmentation results qualitatively. In Figure \ref{fig:Result_inst_seg} (a)$\sim$(d), PolarMask fails to reconstruct complicated parts reliably, such as the bumper of a car or the animals' heads. In contrast, the proposed algorithm represents those boundaries more faithfully. Especially, the proposed algorithm reconstructs the curved parts of object boundaries more accurately. This is because eigencontours consist of a set of curved contours, which are used to represent object contours. However, this combination of eigencontours may create noisy signals, such as wiggling parts in Figure \ref{fig:Result_inst_seg} (e) and (f). Also, both PolarMask and the proposed algorithm assume the star-convexity of object shapes, so they fail to preserve slender and hollow parts of objects, such as the horse legs in Figure \ref{fig:Result_inst_seg} (g).

\section{Conclusions}
In this work, we applied eigencontour descriptors in \cite{park2022} to the instance segmentation task. While the segmentation scheme in \cite{park2022} is based on the ESE-Seg implementation, the proposed network is based on the PolarMask framework. Experimental results demonstrated that the proposed algorithm outperforms PolarMask on COCO2017 and SBD. Also, we analyzed the characteristics of eigencontour descriptors qualitatively.

%%%%%%%%% REFERENCES
{\small
\bibliographystyle{ieee_fullname}
\bibliography{2022_ARXIV_WHPARK}
}

\end{document}